\pdfoutput=1
\documentclass[11pt]{article}

\usepackage[final]{acl}

\usepackage{times}
\usepackage{latexsym}

\usepackage[T1]{fontenc}
\usepackage[utf8]{inputenc}

\usepackage{microtype}

\usepackage{inconsolata}

\usepackage[most]{tcolorbox}
\usepackage{enumerate}
\usepackage{color}
\usepackage{algorithm}
\usepackage[noend]{algpseudocode}
\usepackage{graphicx}
\usepackage{subfigure}
\usepackage{subcaption}
\usepackage{enumitem}
\usepackage{hyperref}
\usepackage{booktabs}
\usepackage{amssymb}
\usepackage{tabularx}

\title{The Real, the Better: Aligning Large Language Models with \\ Online Human Behaviors}

\author{Guanying Jiang\textsuperscript{\textasteriskcentered} \quad Lingyong Yan\thanks{Equal Contribution.} \quad Haibo Shi \quad Dawei Yin\thanks{Corresponding author.}\\
        Baidu Inc. \\
        {\tt \{jiangguanying, yanlingyong, shihaibo\}@baidu.com \quad yindawei@acm.org}}

\begin{document}
\maketitle

\begin{abstract}
Large language model alignment is widely used and studied to avoid LLM producing unhelpful and harmful responses.
However, the lengthy training process and predefined preference bias hinder adaptation to online diverse human preferences.
To this end, this paper proposes an alignment framework, called Reinforcement Learning with Human Behavior (RLHB), to align LLMs by directly leveraging real online human behaviors.
By taking the generative adversarial framework, the generator is trained to respond following expected human behavior; while the discriminator tries to verify whether the triplets of query, response, and human behavior come from real online environments.
Behavior modeling in natural-language form and the multi-model joint training mechanism enable an active and sustainable online alignment.
Experimental results confirm the effectiveness of our proposed methods by both human and automatic evaluations. 
\end{abstract}

\section{Introduction}
Large language models (LLMs) have recently emerged with powerful capabilities and potential for understanding human instructions and generating high-quality answers.
Their impressive intelligence thus promotes many downstream applications, e.g., question answering~\citep{kamalloo-etal-2023-evaluating}, tool learning~\citep{webcpm,toolformer}, and information seeking~\citep{zhu2023large}.
To acquire powerful capabilities, most of them, like InstructGPT~\citep{ouyang2022training} and Llama2~\citep{touvron2023llama}, are first trained over massive language corpora by simply next token prediction learning.
Then they will be fine-tuned over delicately constructed instruction-following datasets, which aims to enhance LLMs to respond to human questions correctly.

\begin{figure}[!t]
  \centering
    \includegraphics[width=0.49\textwidth]{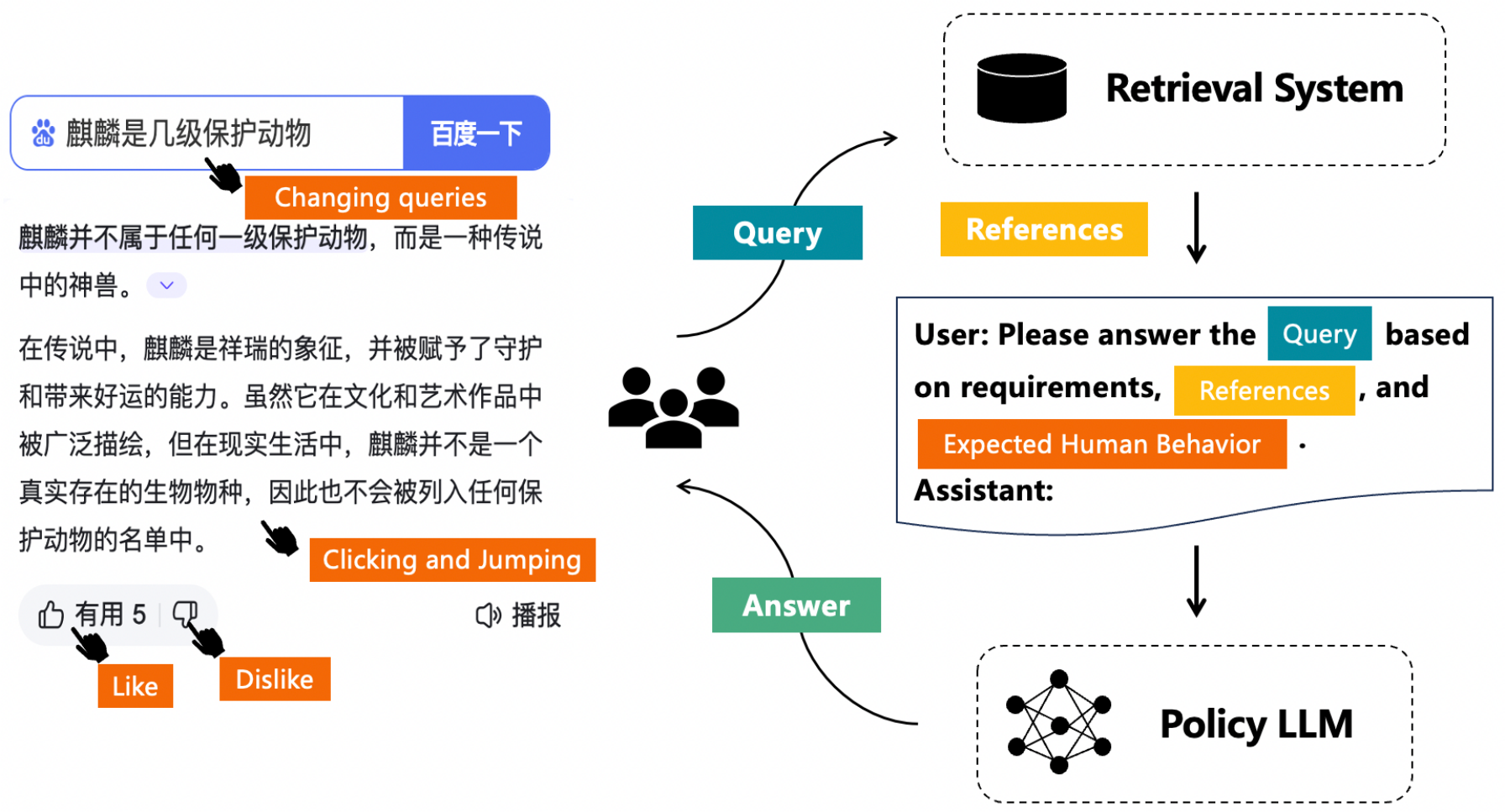}
    \caption{Illustration of collecting online human behaviors in Baidu Search. When a user enters a search query, the answer generated by the LLM can appear at the forefront of the search results.
    Then, the user can interact with the system through various behaviors, such as clicking the contents, giving a like or dislike, or changing the query.}
    \label{fig:model}
\end{figure}

However, fine-tuned LLMs are often observed to produce unexpected or even harmful answers, which is far from human-preferred behavior.
Therefore, most LLMs introduce an alignment phrase to steer the LLM to predefined dimensions, e.g., harmless, helpful, and honest. 
Typically, the LLM alignment methods~\citep{ouyang2022training, touvron2023llama, zheng2023secrets} usually first assign preference signals to model-generated answers from different sources and then train the model based on preference signals via methods like RLHF~\cite{ouyang2022training} and DPO~\cite{rafailov2023direct}.
Specifically, most recent methods usually use an auxiliary reward model to assign preference signals, mainly trained from human-annotated data or human standards for alignment. 
But these approaches always require high labor and time costs and a large amount of inference sampling.
In contrast, some other studies consider AI-assisted annotation or AI-guided feedback as alignment signals to avoid human efforts.
Nevertheless, the feedback distribution from both sources is usually inconsistent with online users, which is time-variant and diverse.

To address the above problems, this paper explores directly leveraging online human behaviors to align LLMs.
To this end, we first demonstrate LLM responses to users for random queries at the Top-1 position of search engine results.
Then, real online human behavior in various dimensions, such as liking and clicking, is collected anonymously.
Later, the behavior signals are processed into numerical or natural-language forms for better LLM alignment with human preferences.

Based on online human behaviors, this paper further proposes a novel LLM alignment framework, \textbf{R}einforcement \textbf{L}earning with \textbf{H}uman \textbf{B}ehaviors (RLHB).
RLHB takes the target LLM as the generator and another auxiliary LLM as the discriminator.
The discriminator takes the query, the generator's response, and the real/fake behavior signal as inputs, and determines whether the <query, response, human behavior> triplets are collected from online environments.
The generator needs to respond to the given query following specified human behavior, and make the <query, response, human behavior> as realistic as possible.
So that, it can confuse the discriminator.
To fully take advantage of LLM's ability to understand and follow natural language, human behaviors, working as condition information, are expressed in natural language form and put into generation instructions.

In this way, the generator and the discriminator are thus trained adversarially.
After convergence, the generator is well-aligned to generate responses that match given human behaviors.
In the inference stage, the well-aligned generator can be directly deployed online, taking the user's query and the most preferred behavior signals as inputs.
Compared to RLHF:
(1) RLHB eliminates annotation requirements and thus can be generalized to various scenarios and applications.
(2) RLHB can continuously learn as human behavior is updated, owing to its multi-model simultaneous training mechanism and behavior modeling in natural-language form.

To verify the effectiveness of RLHB, we conduct a series of experiments and evaluate the performance of RLHB using both human and automatic (i.e., GPT4) evaluation methods.
In summary, our contributions are tri-folds:
\begin{itemize}
\item We propose a novel framework, Reinforcement Learning with Human Behaviors, to align LLMs with online human behaviors. 
\item We construct various experiments to explore how to leverage online human behaviors, from signal combination strategies to signal forms.
\item We verified the model on the online platform, which is evaluated by both humans and GPT4.
\end{itemize}

\section{Related Work}

Aligning large language models with human preferences is first proposed by \citet{stiennon2020learning,ouyang2022training}, which usually adopts reinforcement learning methods\cite{stiennon2020learning,ouyang2022training, lee2023rlaif, bai2022constitutional}, or ranking-based learning methods, such as RRHF \citep{yuan2023rrhf}, DPO \citep{rafailov2023direct}, PRO \citep{song2023preference}, $\Psi$PO \citep{azar2023general}, et al. 
Then, researchers are inspired to study how to align LLMs with different kinds of preference signals, such as human annotations and principles, AI assistance, and online demonstrations.

\paragraph{Human Annotations and Principles.}
In addition to original preference annotation setups in~\citet{ouyang2022training}, some other studies also explore annotating the human preferences in different grains~\cite{wu2023fine}, from fusing sources~\cite{zeng2023diverse, rame2023rewarded}, or through multiple processes~\cite{lightman2023let, uesato2022solving, yuan2023scaling, luo2023wizardmath}.
Besides, \citet{xu2023pinpoint, wang2023shepherd, jin2023data, li2023laffi} tried to directly utilize natural language feedback from the annotators rather than learning a scalar reward.
However, human annotations are usually costly and time-consuming.
Therefore, \citet{bai2022constitutional,anthropic2023ccai,wang2023shepherd} proposed to leverage the meta principles or community feedback to enhance the alignment.
Furthermore, some studies adopt rule-based \citep{openai2023gpt4} and principle-following \citep{sun2023salmon} reward models in their alignment algorithms.

\paragraph{AI Assistance.}
Different from manual feedback or annotations, \citet{chang2023learning} tries to leverage the feedback from other powerful LLMs (e.g., GPT-4) as guidance to align their unaligned LLMs.
Due to the limitation of accessing those powerful LLMs,
\citet{bai2022constitutional, lee2023rlaif, tunstall2023zephyr, yang2023rlcd} propose to first distill the feedback from powerful LLMs to smaller reward models and then use the distilled reward models in alignment practice.
Apart from the AI feedback, the sample quality critique can also be used for LLM alignment.
For example, \citet{shi2023safer} proposed to identify harmful responses and revise them using other LLMs for further fine-tuning and alignment;
while \citet{bai2022constitutional, li2023rain, dong2023raft} iteratively employed self-critique to detect bad responses and revise responses themselves.
Moreover, \citet{dong2023steerlm, hu2023aligning} utilized self-critique rewards as condition information in the generation process, which is similar to our method.

\paragraph{Online Demonstrations.} 
Nevertheless, aligning language models with the aforementioned offline signals could usually cause mis-generalization or distributional shifts, easily resulting in LLM collapses~\cite{casper2023open,ji2023ai}.
Even, the LLM alignment performance is found to be highly sensitive to the noise rate in preference data~\cite{gao2024impact}.
Therefore, \citet{casper2023open, ji2023ai} propose to utilize inverse reinforcement learning (IRL) to model human preferences directly, which infer reward signals by leveraging expert trajectories \citep{ng2000algorithms}. 
However, IRL is often limited to computational cost, demonstration capabilities, and expert efforts when modeling the reward signals~\cite{ho2016generative,ji2023ai,casper2023open}.
Thus, generative adversarial imitation learning (GAIL)~\cite{ho2016generative} is proposed to fuse IRL into Generative Adversarial Nets (GAN) \citep{goodfellow2014generative}, where the actor tries to generate approaching high rewards, and the discriminator assesses whether the trajectory is expert-generated.

\section{From Human Feedback To Human Behaviors}
\label{humandata}
Leveraging online anonymous human behaviors to improve content quality and user experience has been studied for a long period in Information Retrieval ~\cite{joachims2017accurately, mitra2018introduction, huang2020embedding} and Recommendation Systems~\cite{hu2008collaborative, liu2010unifying, zhao2018explicit, xie2021deep, wu2022feedrec}.
The leveraged human behaviors can be usually divided into three types: the explicit behaviors~\cite{huang2020embedding}, the implicit behaviors ~\cite{hu2008collaborative, joachims2017accurately}, and the fused ones~\cite{liu2010unifying, zhao2018explicit, xie2021deep, wu2022feedrec}.
Explicit behavior refers to users' proactive feedback behavior.
For instance, user preferences can be directly collected by counting the times they click the \texttt{Like} or \texttt{Dislike} button.
Other interactive patterns are used like \texttt{Sharing} (demonstrates users' preference for the content by actively distributing it), \texttt{Commenting} (illustrates the inclination to participate in topic discussion actively), and so on.
Implicit behavior usually involves indirect and non-perceived interactions, like \texttt{Page Views} (abbreviated as PV, representing the number of times a query is searched and exposed), \texttt{Clicks} (denote the number of times an answer is clicked, reflecting the level of interest or engagement a user has), \texttt{Dwell Time} (also reflects users' interests and concerns in the presented content), \texttt{Switching to Similar Queries} (means the answer does not satisfy the user's full intent), and so on.

Recently, more and more web search engines have been committed to fulfilling user requirements relying on the LLM to directly return satisfied responses, and present them on the forefront page (see Figure~\ref{fig:model} as an example). 
Through multiple user interactions, such as clicking, liking, changing queries, etc., the system continuously updates the generated answers until the customer is satisfied. 
In this process, the system accumulates amounts of real interactive trajectories. Compared with manually annotated predefined preference data, real online interactive behaviors are more consistent with users’ habits and preferences.

In this work, to simplify, we fuse four types of representative explicit and implicit indicators to describe users' preferences, i.e. Page Views, Clicks, Likes, and Dislikes.
The first three are positive indicators and the last is negative.
We smooth indicators by $\log(1+x)$ and discretize each into $N$ equal parts.
During RL training, we perform reward shaping $\left(\texttt{Likes} - \texttt{Dislikes}\right) / \left(\texttt{PV} + \texttt{Clicks}\right)$, combining multi-head rewards into a holistic scalar.

\section{LLM Alignment with Human Behaviors}
Based on the above human behaviors, we introduce how to leverage them for LLM alignments in this section.
We propose two alignment methods using human behaviors, denoted as RLHBC and RLHB.

\subsection{Problem Definition}
LLM alignment with online human behaviors can be formulated as a Markov Decision Process (MDP) problem, denoted as a tuple $(\mathcal{S}, \mathcal{A}, \mathcal{P}, \mathcal{R}, \gamma)$.
We consider human interactions and behaviors as the environment $E$. At every time step $t$, the LLM agent observes the current state $s_t \in \mathcal{S}$ from $E$ (the query triggered by the customer) and takes actions $a_t \in \mathcal{A}$ according to a policy $\pi: \mathcal{S} \mapsto p(\mathcal{A})$ that maps the states to a probability distribution over the actions. The agent’s action is to generate tokens till the end of the content. Then, the agent will receive a reward $r_t = \mathcal{R}(s_t, a_t)$ from the trained reward model, and a new state $s_{t+1} \in \mathcal{S}$ from the environment $E$. The return of the interactive trajectory $\tau=\{s_1, a_1, \dots, s_T, a_T\}$ is the cumulative $\gamma$-discounted rewards, i.e. $R(\tau) = \Sigma_{t=1}^{T}{\gamma^tr_t}$, where $T$ is the horizon of an episode. RL aims to optimize the policy $\pi$ by maximizing the expected returns from the initial state.

\subsection{A Naive Method}
\label{sec:class}
Given human behaviors towards <question, LLM response> pairs, one naive method is to directly train feedback simulators to predict user behaviors $b\in \mathcal{B}$ based on the query $s\in \mathcal{S}$ and the LLM response $a\in \mathcal{A}$.
Then, the trained simulators can be used as reward models in the RLHF framework to align LLMs with real human behaviors. 

In practice, however, collecting sufficient human behaviors related to different answers to the same questions is usually difficult.
For example, we may only have one chance to provide the LLM response if users search for some queries that usually never be searched again in a web search engine.
Therefore, we propose to build a multi-head pointwise classifier with cross-entropy loss as follows:
\begin{equation}
\label{eq:rm-class}
\text{CE}(\mathbf{\hat{b}}, \mathbf{b}) = -\frac{1}{N} \sum_{i=1}^{N} \sum_{j=1}^{C} b_{ij} \log(\hat{b}_{ij})
\end{equation}

After that, the multi-head pointwise classifier is used as the reward model to align LLMs via RLHF algorithm.
We thus call it \textbf{R}einforcement \textbf{L}earning with \textbf{H}uman \textbf{B}ehavior through \textbf{C}lassifier, \textbf{RLHBC}.

\subsection{Reinforcement Learning with Human Behaviors}
\label{cgail}
Inverse reinforcement learning (IRL) is a popular alternative to align target models with online expert demonstrations, with the advantage of not interacting with experts during training. It enhances sampling and training efficiency, compared to reinforcement learning and imitation learning.
Generative Adversarial Imitation Learning (GAIL)~\citep{ho2016generative} can further bypass the intermediate step of recovering the unknown reward model in IRL. It directly optimizes the policy, using a GAN discriminator trained by expert demonstrations to provide the action-value function,
\begin{equation}
\label{eq:pre1}
\begin{aligned}
 Q(s, a)=\mathbb{E}_{ \left(s, a\right) \in \mathcal{M}}\left[\log \left(D_{\omega}(s, a)\right)\right],
\end{aligned}
\end{equation}
where the expert and generated demonstrations are denoted as $\mathcal{M}_e$ and $\mathcal{M}_g$. The initial policy actor and discriminator parameters are $\theta_0$ and $\omega_0$.
The discriminator $D$ can be updated by maximizing
\begin{equation}
\label{eq:disc_1}
\begin{aligned}
&\mathbb{E}_{\left(s, a\right)\in \mathcal{M}_e}\left[\log(D_{\omega}\left(s, a\right)) \right] + \\ &\mathbb{E}_{\left(s, a\right)\in \mathcal{M}_g}\left[1 - \log(D_{\omega}\left(s, a\right)) \right].
\end{aligned}
\end{equation}
The discriminator evaluates whether the interaction trajectory comes from expert demonstrations; while the generator is promoted to generate expert-like high-quality content.

In industrial scenarios, online demonstrations with strong positive feedback are considered expert or golden demonstrations. However, strong feedback samples are usually sparse and imbalanced. 
For example, most search items may receive at most one view or click, but others can attract rich and massive interactions by most users.

\begin{figure}[t]
  \centering
    \includegraphics[width=0.49\textwidth]{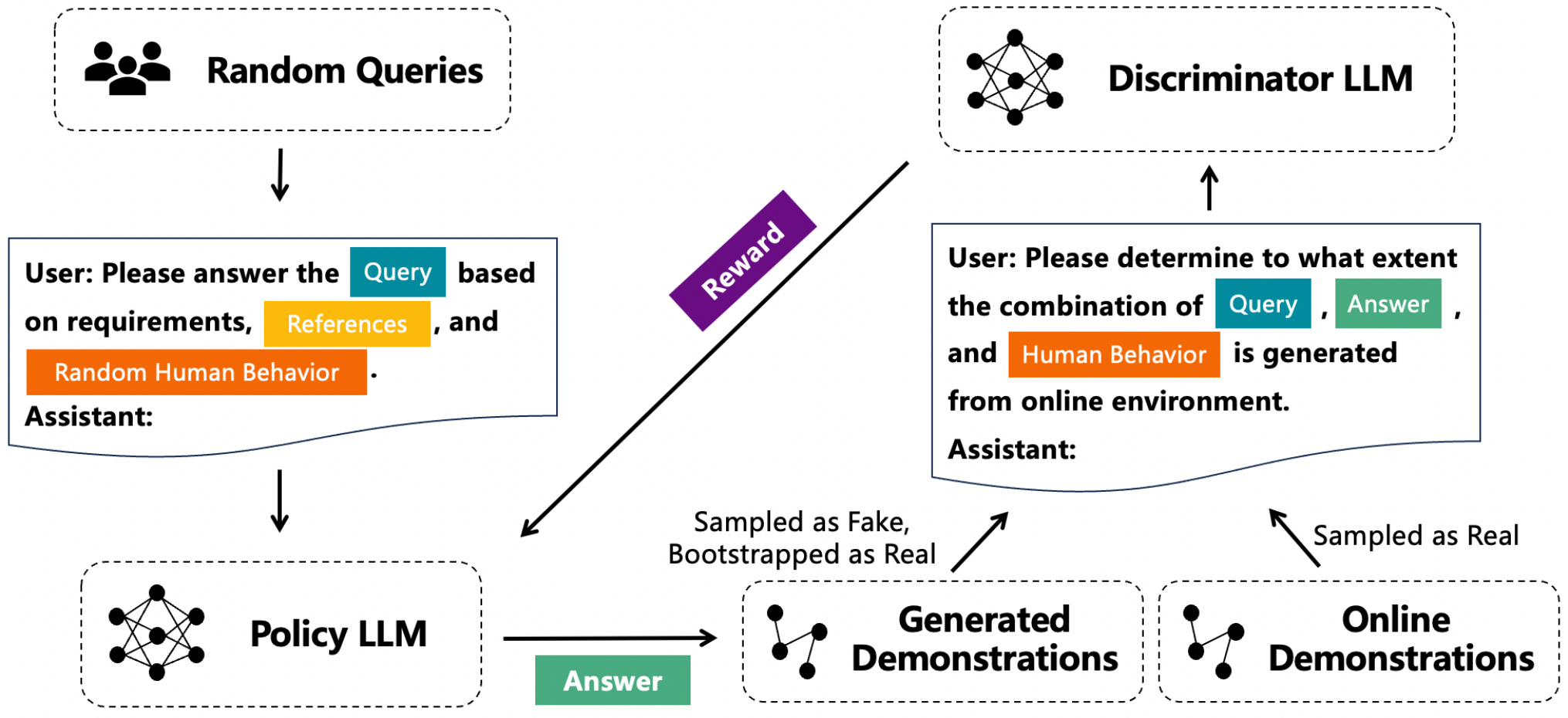}
    \caption{The training process of RLHB. }
    \label{fig:model-train}
\end{figure}

Regarding the above problem, we propose a novel alignment method, named \textbf{R}einforcement \textbf{L}earning with \textbf{H}uman \textbf{B}ehaviors (\textbf{RLHB}, see Figure~\ref{fig:model-train} for illustration), fully utilizing online demonstrations, inspired by Decision Transformer \citep{chen2021decision}. The discriminator here is defined to verify whether the query-answer <$s$, $a$> pair under the given feedback $b$ comes from real online demonstrations, rather than expert demonstrations.
\begin{equation}
\label{eq:disc_2}
\begin{aligned}
&\mathbb{E}_{(s_{t}, a_{t}, b_{t})\in \mathcal{M}_e}\left[\log(D(s_{t}, a_{t}; b_{t})) \right] + \\ &\mathbb{E}_{(s_{t}, a_{t}, b_{t})\in \mathcal{M}_g}\left[1 - \log(D(s_{t}, a_{t}; b_{t})) \right]
\end{aligned}
\end{equation}
The LLM generator, when receiving a query $s$, is supposed to give a response $a$ following the expected feedback $b$, that is, $\pi\left(a_t | s_t, b_t \right)$, regardless of positive or negative feedback. When implemented in an online environment, the policy is required to comply with preferred feedback.

We update the generator using the objective function with clipped surrogate following RLHF:
\begin{equation}
\label{eq:disc_3}
\mathcal{L}(\theta)=\mathbb{E}_{t}\left[\min\left( \ell_t \hat{A}_{t}, \operatorname{clip}\left(\ell_{t}, 1-\epsilon, 1+\epsilon\right) \hat{A}_{t}\right)\right],
\end{equation}
where the ratio of the new policy over the old is
\begin{equation}
\label{eq:l_theta}
\ell_t = \frac{\pi_{\theta}\left(a_{t} \mid s_{t}, b_{t}\right)}{\pi_{\theta_{\mathrm{old}}}\left(a_{t} \mid s_{t}, b_{t}\right)},
\end{equation}
and the advantages are calculated following the generalized advantage estimator (GAE):
\begin{equation}
\label{eq:disc_5}
\hat{A}_{t}=\sum_{l=0}^{\infty}(\gamma \lambda)^{l} \delta_{t+l},
\end{equation}
\begin{equation}
\label{eq:disc_6}
\delta_{t}=r_{t}+\gamma V_{\phi}\left(s_{t+1}; b_{t+1}\right)-V_{\phi}\left(s_{t}; b_{t}\right),
\end{equation}
\begin{equation}
\label{eq:disc_8}
\hat{R}_t = \hat{A}_{t} + V_{\phi}\left(s_{t}; b_{t}\right).
\end{equation}
The 'rewards' $r_t$ is provided by the discriminator $D(s_{t}, a_{t}; b_{t})$, and $\hat{R}_t$ denotes expected returns. 
Note that $r_t$ is combined with the KL divergence penalties per token to prevent deviation from the initial model.
\begin{equation}
\label{eq:rm_kl_penalty}
r_t - \eta\mathrm{KL}\left(\pi_{{\theta}_{\mathrm{old}}}(a_t\mid s_t, b_t), \pi_{\theta}(a_t\mid s_t, b_t)\right)
\end{equation}
The critic, with parameters $\phi$, providing token-level state-values, is updated by
\begin{equation}
\label{eq:disc_7}
\mathcal{L}(\phi)=\mathbb{E}_{t}\left[ \| V_{\phi}\left(s_{t}; b_{t}\right) - \hat{R}_t  \| ^2 \right].
\end{equation}
The parameters of the policy $\theta$, critic $\phi$, and discriminator $\omega$ models are iteratively updated in sequence during training.

\paragraph{Bootstrap Enhancement}
Since RLHB introduces simultaneous training of models (i.e. the actor, critic, and discriminator), the efficiency and balance of sampling are crucial to each. 
The outputs of the critic and discriminator are both scalar values, which are relatively easy to learn; in contrast, the convergence of the actor is more difficult owing to larger and more uncertain action space. 
To prevent the discriminator from overfitting and make it more robust, we bootstrap a certain proportion $\kappa$ of highly-rewarded demonstrations as 'fake' online demonstrations for discriminator updating. 

\paragraph{Behavior in Natural-Language Form}
Furthermore, we change the form of behavioral data modeling.
In RLHF-type methods, data processes in pairwise or numerical forms are necessary before reward modeling and RL alignment, breaking the training process into parts and bringing instability in signal models.
However, LLM has proven its ability to understand and express natural language like humans, including text containing data and statistical results.
Therefore, under the settings of RLHB, we instead utilize the intrinsic properties of LLM to describe human behaviors in the form of natural language and put it into instructions.

\section{Experiments}

In this section, we conduct several experiments to assess the effectiveness of our methods, which aims to answer the following two questions:
\begin{enumerate}[label={Q\arabic*}]
    \item Whether the generative model can be directly aligned on real online human behaviors? \label{issue-one}
    \item Whether the predefined preferences alignment can be further improved by online alignment? \label{issue-two}
\end{enumerate}

\subsection{Data}
\label{cmdata}

\paragraph{Human Behavior Data.} Online human behavior data is collected from Baidu Search.
We uniformly sample real demonstrations from online environments based on indicators discussed in Section~\ref{humandata}. 
We finally obtained around 100k <query, answer, feedback> triplets for experiments.

\paragraph{Human Preference Data.} We also collect manual preference data for reward modeling, following InstructGPT~\citep{ouyang2022training}. 
We sample random queries and generate multiple answers for each using LLMs for internal usage.
Then, every two different answers to the same query are annotated with preference labels by experienced annotators \footnote{The annotation consistency confidence level is over 80\%}.

\subsection{Models}
\label{fixedmodels}

\paragraph{Baseline (SFT).} We utilize an internal LLM (13B) as the backbone and set its default version as the baseline, fine-tuned well with millions of platform-owned data.

\paragraph{Reward Model (RM).} RM is trained on human preference data with pairwise ranking loss in Eq.~\ref{eq:rmloss}. 
$\psi$ denotes the RM parameters, and $a_{w}$ is the preferred answer for each pair. 
\begin{equation}
\label{eq:rmloss}
\mathcal{L}(\psi) = - \log \sigma\left(r_{\psi}\left(s, a_{w}\right)-r_{\psi}\left(s, a_{l}\right)\right) 
\end{equation}

\paragraph{Classifier Model (CM).} 
Following setups in Section~\ref{sec:class} and \ref{cmdata}, the classifier in RLHBC fits human behavior data in numerical form.

\paragraph{Discriminator Model (DM).}
We also warm up a discriminator for ablation experiments. It is trained using online human behavior samples as real, and random replacements of their feedback as fake.

\subsection{Evaluation Metrics}
Three types of metrics are used for evaluation.
First, as GPT4~\citep{openai2023gpt4} is widely used as a reference to evaluate the quality of generated text~\citep{wang2023shepherd}, we follow these studies to prompt GPT4~\footnote{GPT4-Turbo is used in our evaluation.} to rank each pair of responses to the same query from different models and then summarize them into Win-Tie-Loss (WTL) results.
Moreover, we adopt two human evaluation systems in Baidu Search to provide fine-grained assessments, i.e. Quality Score and Satisfaction Score. We recruit annotators, well-educated and experienced in the labeling criteria, to label these scores for each pair of queries and LLM responses. 
Additionally, following other studies on RL-based alignment methods~\citep{zheng2023secrets}, we utilize RL indicators to evaluate the stability of model training and LLM metrics to assess the LLM capabilities. 

\paragraph{Quality Score.} It focuses on content quality with four levels. 
\textbf{Bad} indicates text with low quality, mistakes, irrelevant or false contents; 
\textbf{Medium} means redundant texts or those that do not solve the major need; 
\textbf{Good} denotes the text can meet the need with the lack of content richness, relevance, authoritativeness, or necessary multi-modal information; 
\textbf{Excellent} represents the generated text fully and accurately meets needs and adds in-depth and extended content. Based on Quality Scores, pairwise Win-Tie-Loss can also be provided.

\paragraph{Satisfaction Score.} It is a three-level score to measure user satisfaction with given responses to their search queries.
\textbf{Unsatisfied} mainly indicates content quality problems, such as irrelevance, mistakes, duplications, hallucinations, partial satisfaction of the user's need, and some other defects; 
\textbf{Partially Satisfied} means the answers meet main requirements but in an unsatisfied display form, such as under-listing, over-listing, excessive length, inappropriate integration of multi-modal contents, etc.; 
\textbf{Satisfied} demonstrates that satisfied or even in-depth content is integrated into a proper format and length, without any wrong information. 

\begin{figure*}[t]
\centering
\subfigure[GPT-4 evaluation over quality WTL.\label{fig:gsb_all:a}]{\includegraphics[width=3.1in]{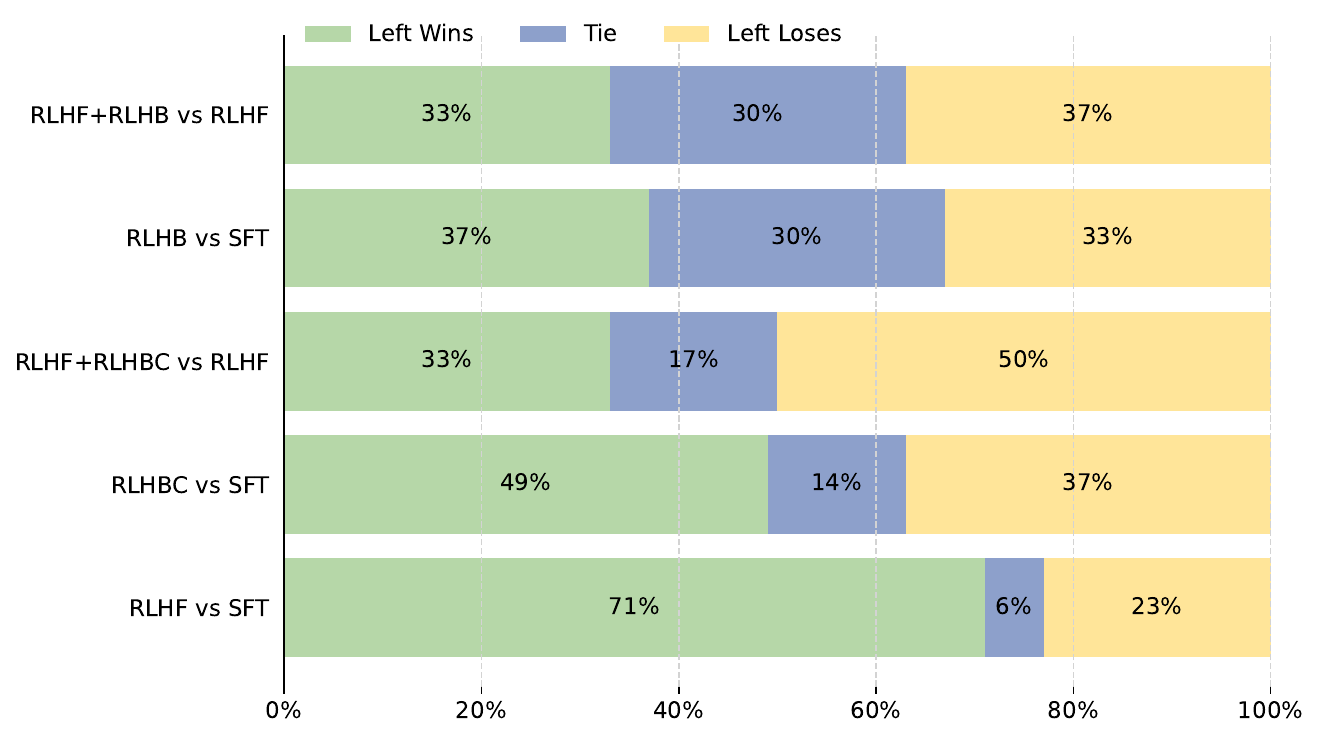}}
\subfigure[Human evaluation over quality WTL.\label{fig:gsb_all:b}]{\includegraphics[width=3.1in]{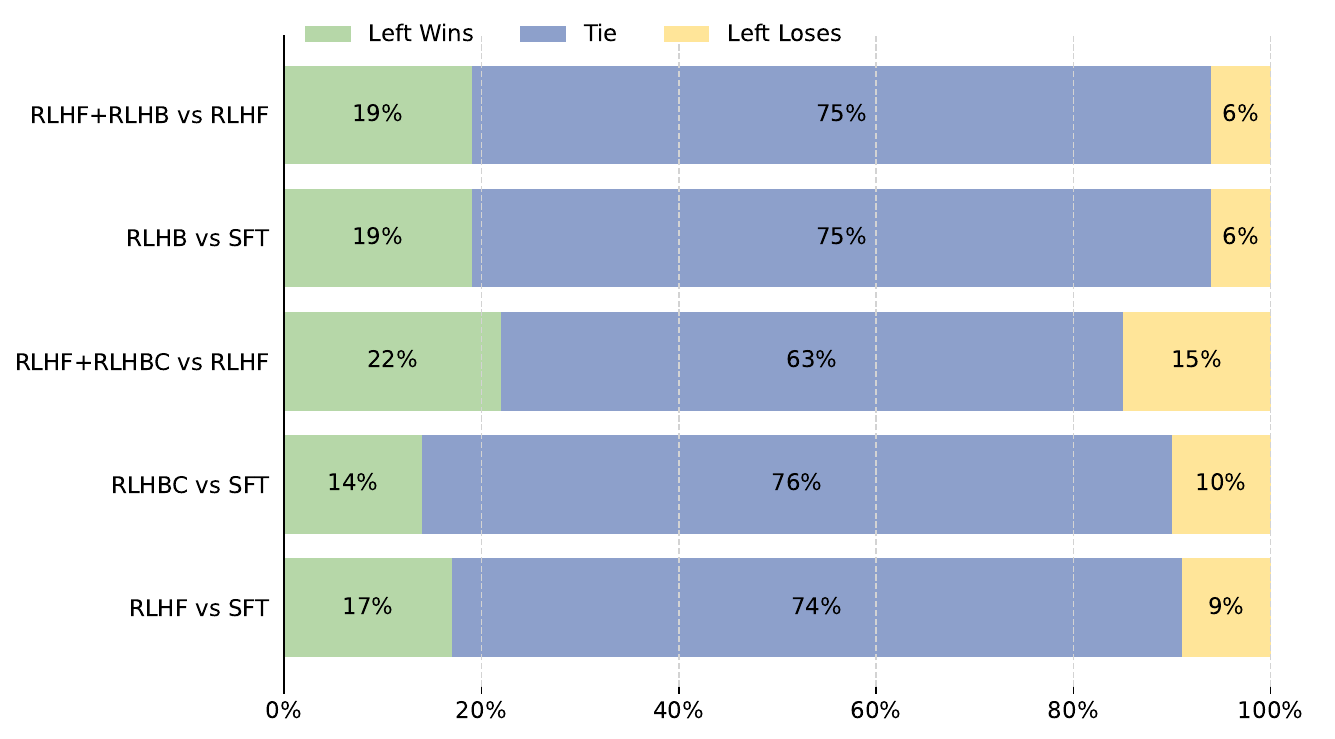}}\\
\subfigure[Human evaluation over quality score.\label{fig:gsb_all:c}]{\includegraphics[width=3.1in]{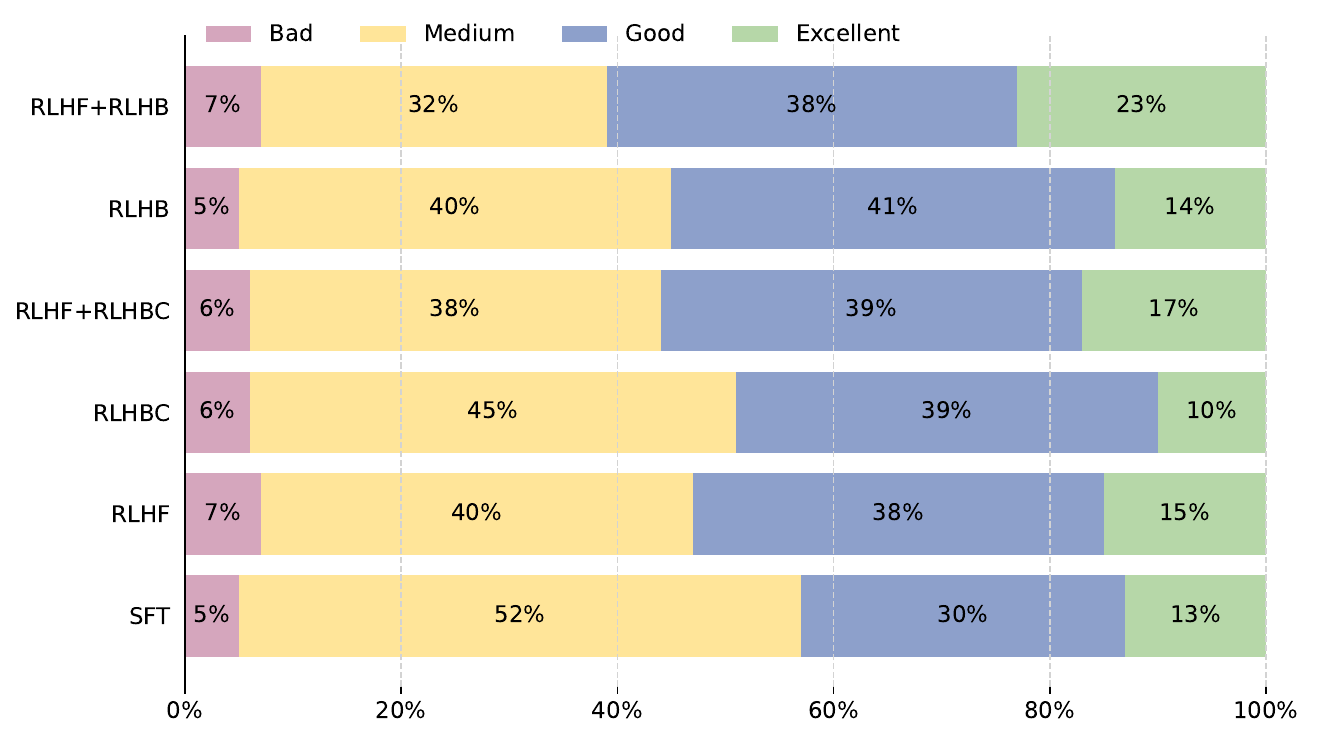}}
\subfigure[Human evaluation over satisfaction score.\label{fig:gsb_all:d}]{\includegraphics[width=3.1in]{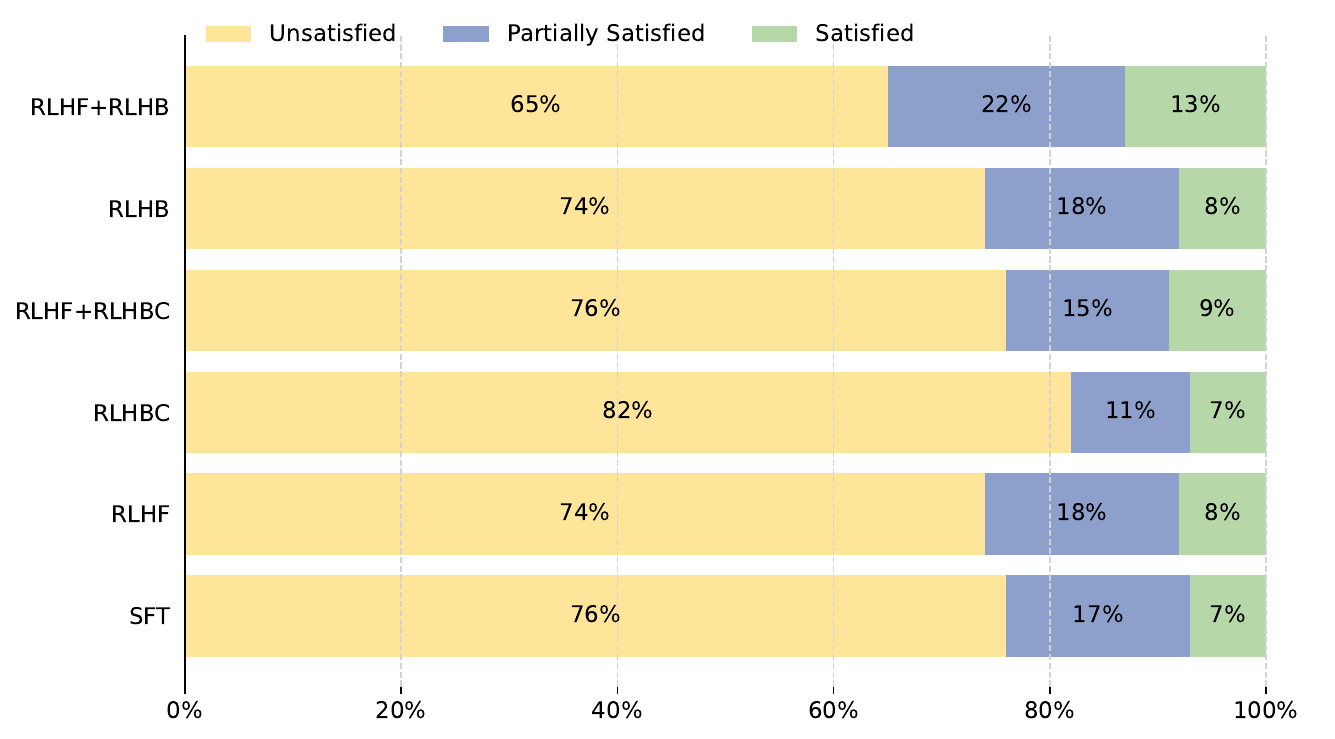}}
\caption{GPT4 and Human Evaluations. 
Even though the results differ to varying degrees, most confirm the feasibility of the proposed questions~\ref{issue-one} and ~\ref{issue-two}, especially from the perspective of RLHB and RLHF + RLHB.}
\label{fig:gsb_all}
\end{figure*}

\subsection{Other Setups}
For~\ref{issue-one}, we trained RLHF, RLHBC, and RLHB from the baseline to see which produces more alignment gains for the unaligned SFT. The critic models for RLHF and RLHBC are started from RM and CM; while the critic and discriminator models for RLHB are initialized from the baseline.
For~\ref{issue-two}, we further train RLHBC and RLHB based on RLHF to see if additional improvements can be made, denoted as RLHF + RLHBC and RLHF + RLHB.

\section{Experimental Results}
\subsection{GPT4 Evaluation}
We obtain generated responses to random queries and combine them into pairs. 
Then, we utilize evaluation prompts to ask GPT4 to select a better response. 
Note that we combine each pair in forward and reverse orders and ask GPT4 to rank them twice. 
The results are shown in Figure~\ref{fig:gsb_all:a}.

In the quantity of Win-Tie-Loss, the RLHBC and RLHB perform better than their baseline SFT; while the RLHF + RLHBC and RLHF + RLHB are worse than their baseline RLHF.
However, based on the Sign Test, the performance of these four groups is statistically equal.
Moreover, it is noteworthy that, compared to SFT, RLHF shows significant improvement, with a win rate of 71\%.
In brief, from the perspective of GPT4, regardless of the baselines, the improvement brought by online interactive alignment is limited and may even weaken the performance in some cases.

However, studies found that the limitations and knowledge barriers of GPT4 may impair its reliability of results \citep{wang2023shepherd}. Thus, we further leverage manual evaluation systems to provide a comprehensive assessment.

\begin{figure*}[t]
\centering
\subfigure{\includegraphics[width=2.0in]{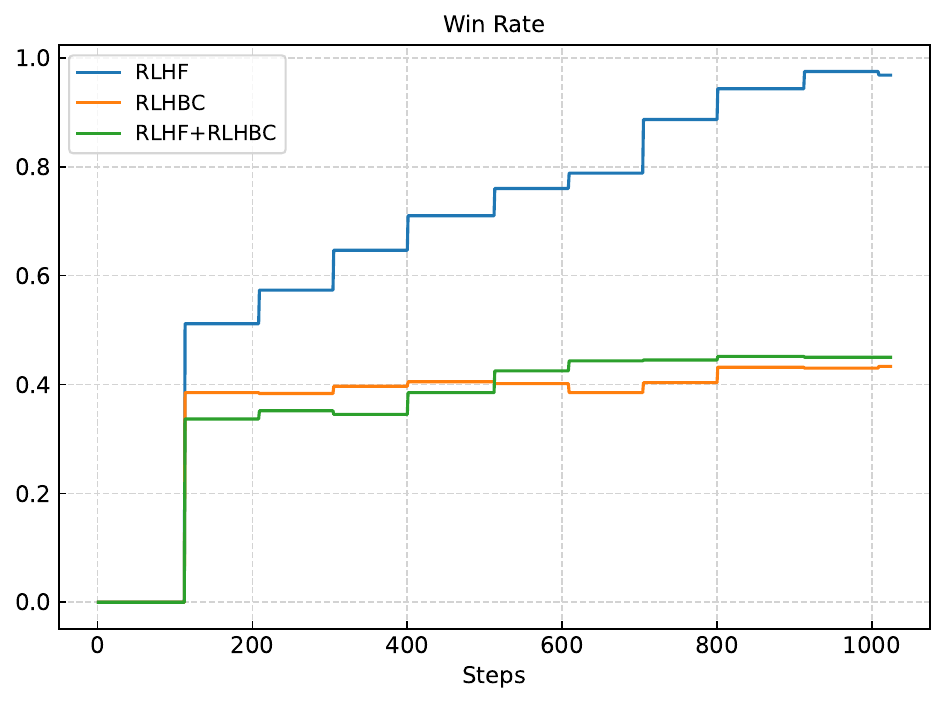}} 
\subfigure{\includegraphics[width=2.0in]{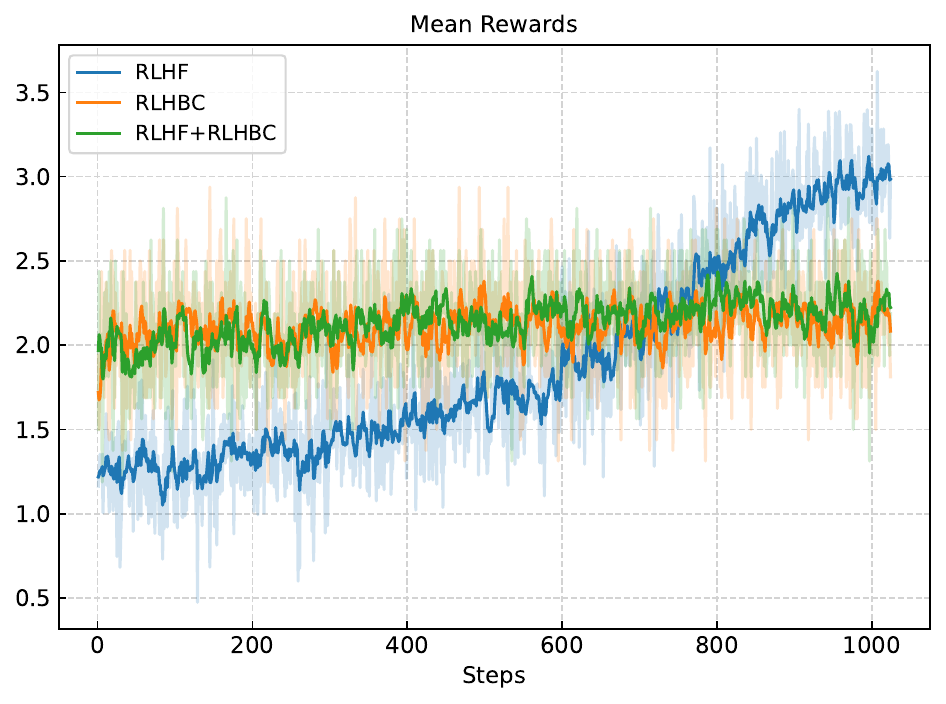}}
\caption{Win Rates and Mean Rewards for RLHF, RLHBC, and RLHF + RLHBC. 
Compared with CM of RLHBC models, RM of RLHF is much easier to guide the model to achieve preference learning and model convergence.}
\label{fig:rlhbc}
\end{figure*}

\begin{figure*}[t]
\centering
\subfigure{\includegraphics[width=2.0in]{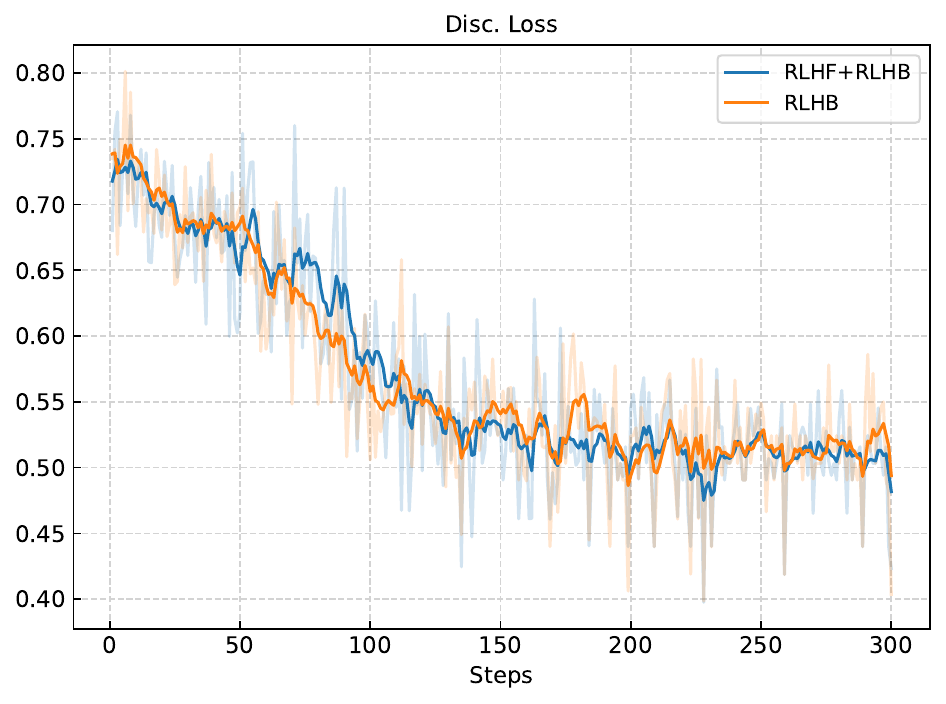}} 
\subfigure{\includegraphics[width=2.0in]{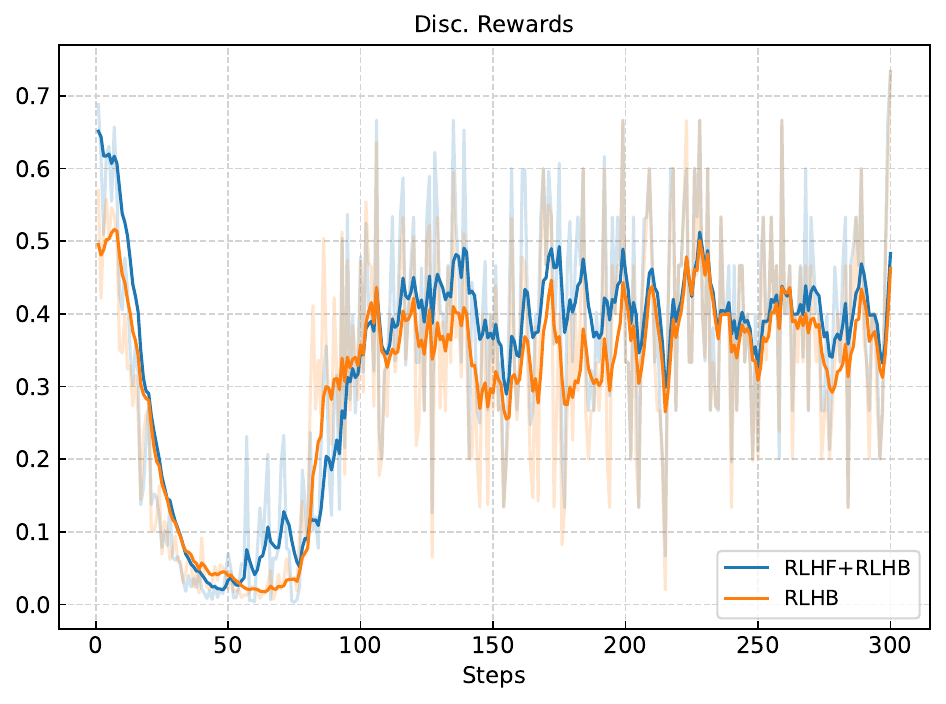}}
\subfigure{\includegraphics[width=2.0in]{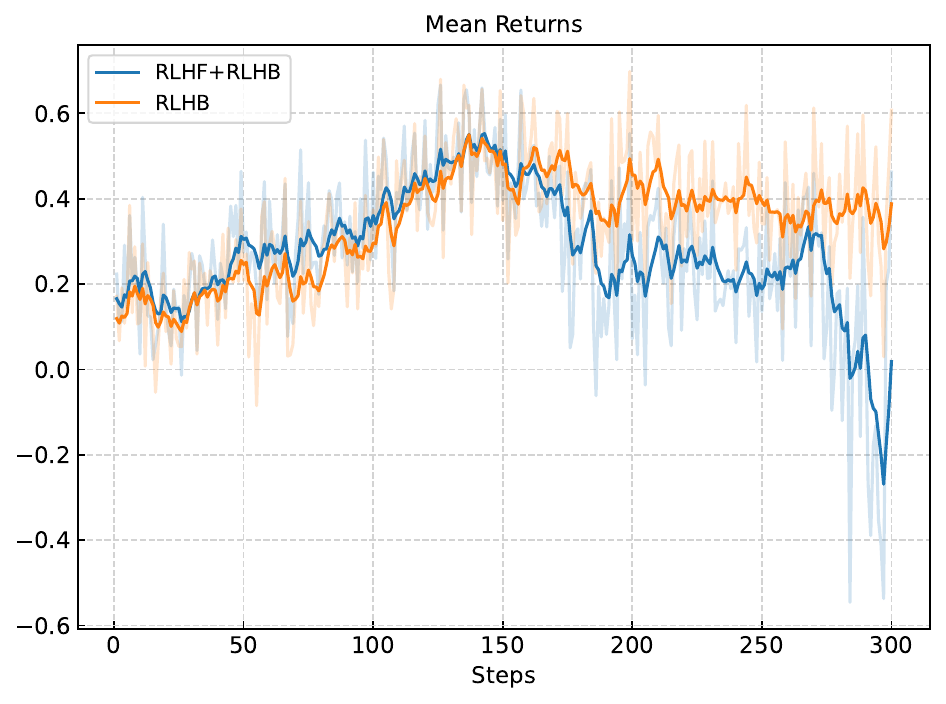}}
\caption{Discriminator Loss, Discriminator Rewards, and Mean Returns for RLHB and RLHF + RLHB.
Different from before, the rewards in RLHB will not continue to grow, but will eventually converge to a confusion state, close to 0.5, instead.
That means the policy generation ability can already confuse the real with the fake,  though this instability may lower the expected returns.
}
\label{fig:rlhb}
\end{figure*}

\begin{figure*}[t]
\centering
\subfigure[\label{fig:abl-disc:a}]{\includegraphics[width=2.0in]{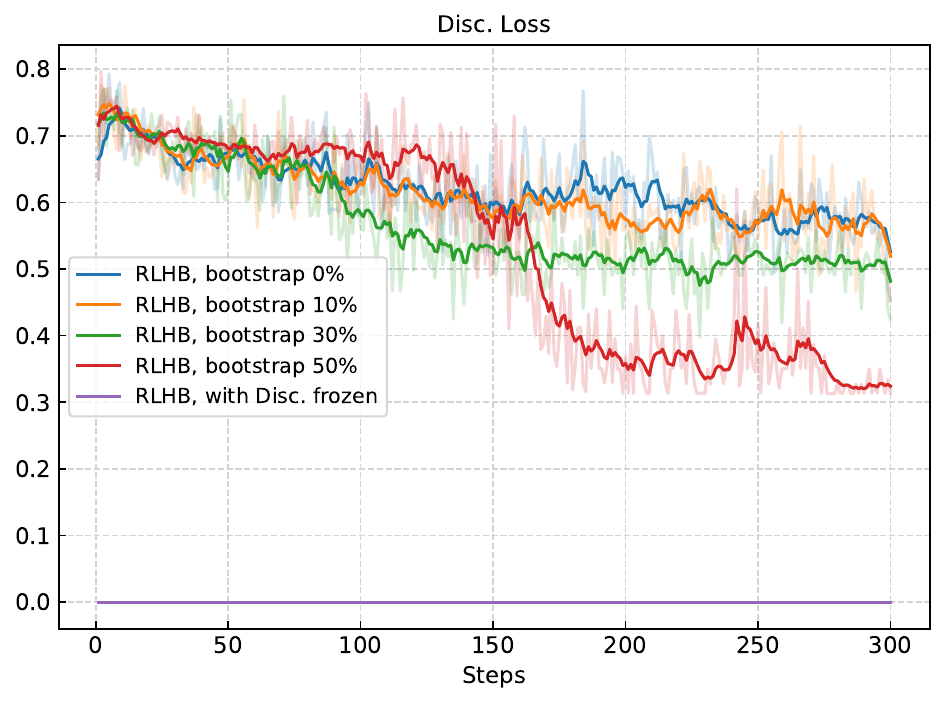}}
\subfigure[\label{fig:abl-disc:b}]{\includegraphics[width=2.0in]{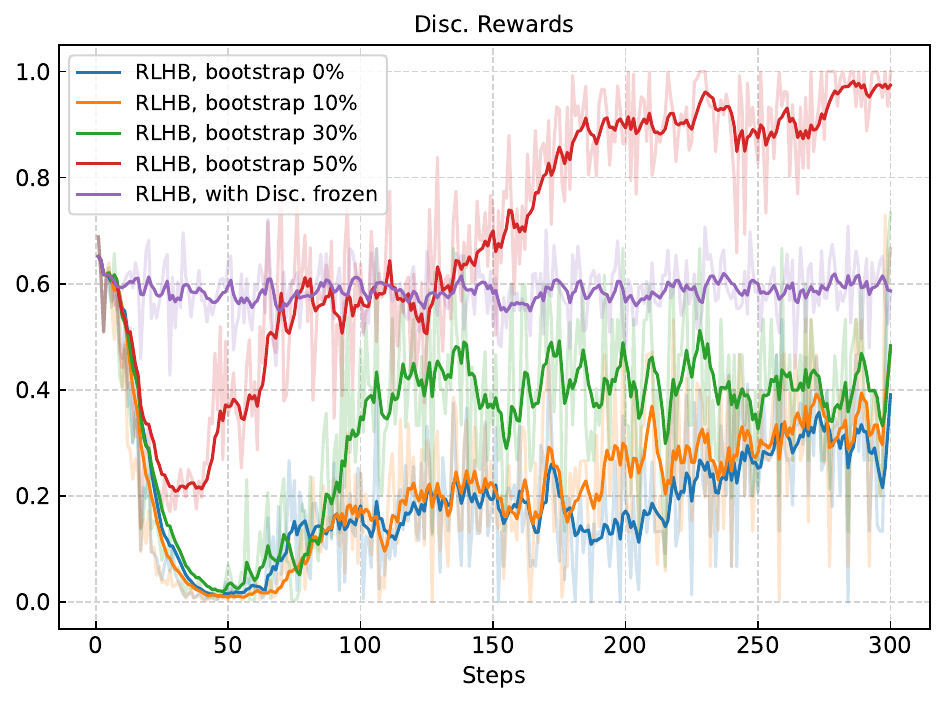}}
\subfigure[\label{fig:abl-disc:c}]{\includegraphics[width=2.0in]{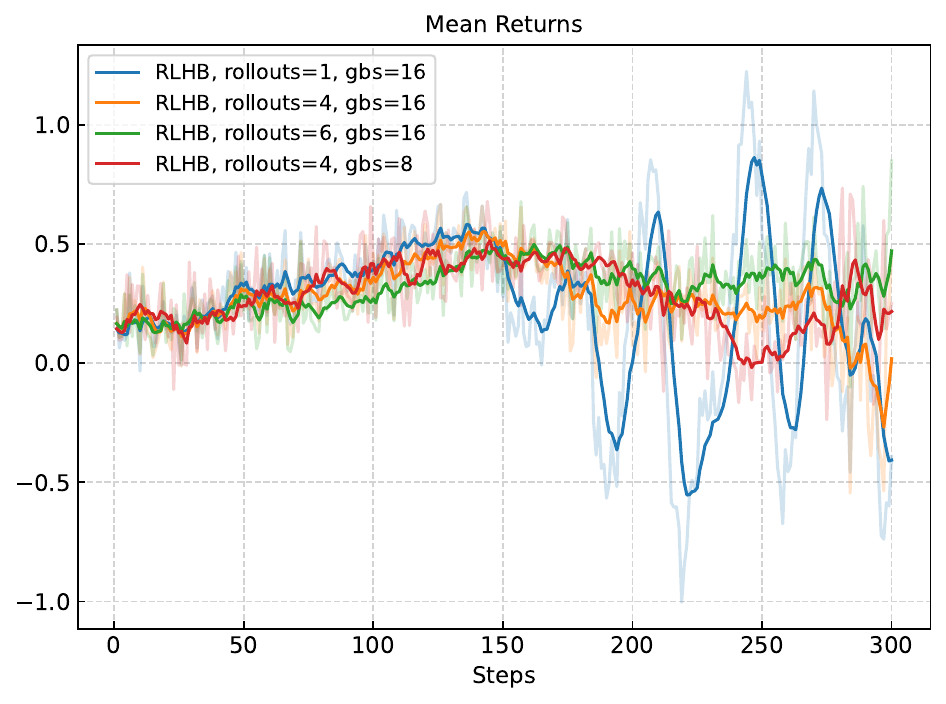}}
\caption{The Ablation Experiments for RLHB.}
\label{fig:abl-disc}
\end{figure*}

\subsection{Human Evaluation}
Figure~\ref{fig:gsb_all:c} demonstrates human evaluation results by Quality Scores, and we mainly focus on Excellent and Good improvements.
First, all candidates show improvements over the baseline SFT, ranging from 6\% to 18\%. 
The performance of RLHB and RLHF is basically the same, indicating question~\ref{issue-one} is feasible.
Yet, RLHBC is not better than RLHF, which may result from the Classifier being sensitive to the training data distribution.
Furthermore, RLHF + RLHBC and RLHF + RLHB achieve improvements of 3\% to 7\% over their baseline RLHF. 
It indicates that question~\ref{issue-two} is also workable, enabling further gains from online interactive alignment based on predefined preference alignment. 
On the whole, RLHF + RLHB is the only one that achieves 60\% high-quality ratio, satisfying users' requirements without hallucinations or defects.

Besides, we report Win-Tie-Loss results according to Quality Scores in Figure~\ref{fig:gsb_all:b}.
Different from GPT4's results, the performance of RLHF and SFT is statistically equal here. 
Additionally, with a rate of Win-Loss 19\%:6\%, RLHF + RLHB and RLHB present significant improvements from RLHF and SFT, respectively. 
It confirms the feasibility of the two questions again.
On the contrary, the enhancement of RLHF + RLHBC and RLHBC over RLHF and SFT is just near statistical significance.
Another problem occurs to RLHF + RLHBC. 
When the win rate increases to 22\%, the loss rate also rises to 15\%, both the highest among all comparisons.
Therefore, RLHB-based models are more promising in industrial applications.

Figure~\ref{fig:gsb_all:d} displays the outcomes of Satisfaction Scores, and we mainly focus on the percentages of Satisfied and Partially Satisfied.
Firstly, RLHB and RLHF perform similarly; while RLHF + RLHB further improves RLHF by 9\%. These once again confirm the two questions are feasible.
However, both RLHF + RLHBC and RLHBC are worse than their baselines RLHF and SFT, respectively. It reflects the stability issues of Classifier-based alignment models.
Interestingly, on Satisfaction dimensions, the effects of RLHF and SFT are still the same, which is quite different from GPT4's results.

\subsection{Model Training Metrics}
In this subsection, we study the influences of hyper-parameter setups on model performances.
The training metrics for RLHF, RLHBC, and RLHF + RLHBC are presented in Figure~\ref{fig:rlhbc}. 
It is worth noting that the win rates by RM and mean rewards for RLHF present obvious upward trends, approaching 0.95 and 3.0 respectively.
In contrast, for RLHBC and RLHF + RLHBC, the win rates fluctuate around 0.4, and the mean rewards reach 2.25 only.
Compared with other methods, RLHF achieves better convergence under the incentive of RM.

The training metrics for RLHB and RLHF + RLHB are shown in Figure~\ref{fig:rlhb}.
We can see that the discriminator loss tends to converge stably.
However, the rewards of discriminators show different trends. 
Starting around 0.5, those reward scores decline sharply while the discriminative abilities increase. 
As the policy is optimized, the rewards decline to around 0.4 in contrast, showing the discriminator is confused to some extent.
Yet, the instability of discriminators affects the critic model fitting, leading to fluctuations in expected returns.

\subsection{Alabation Study}
In Figure~\ref{fig:abl-disc}, we conduct several abalation experiments.
As for the discriminator, we set $\kappa$ to different values ranging from  0\% to 50\%, to study the model sensitivity to the proportion of fake-real data, as discussed in Section~\ref{cgail}.
In addition, we also experiment on the trained discriminator in Section~\ref{fixedmodels} and freeze its parameters during RLHB training, to see if the discriminator can be trained and used in the way of the reward model.
As for the actor (i.e., the LLM policy agent), we compare different numbers of rollouts and batch sizes, set to 4 and 16 by default respectively, to see the impact of the sampling scale.

For the discriminator, as shown in Figure~\ref{fig:abl-disc:a} and \ref{fig:abl-disc:b}, when the bootstrap proportion $\kappa$ reaches 50\%, the discriminator collapses sharply, with the rewards approach almost 1.0. 
Moreover, the frozen discriminator tends to predict the generated samples at a stable score of around 0.6.
It cannot be observed that the impact of improved policy generation capabilities on the discriminator.
Thus, without the adversarial mechanism, the discriminator cannot work like RM.
For the actor, see Figure~\ref{fig:abl-disc:c}, with rollouts increasing from 1 to 4, and 6, the stability can be significantly improved, especially from mean returns.
By changing the batch size 16 to 8, the variance in training effect is not obvious, except for a slightly larger volatility.

\section{Conclusions and Discussion}
In this work, we explore aligning LLM with the behavioral preferences from online users.
We propose two alignment methods, RLHBC and RLHB, and consider online behavior in both natural language form and scalar form.
Numerous experiments show that:
On the one hand, the LLM alignment based on online human behavior can approximate the alignment based on offline annotated preferences;
On the other hand, it can also be concluded that online behavior alignment can further enhance the LLM aligned with offline preferences.

RLHB exhibits numerous advantages compared to RLHF and RLHBC.
Whether it is RLHF or RLHBC, the reward model or classifier model must be trained with preprocessed scalar-form human behavior before RL alignment, thus preventing the model from actively and continuously learning online.
On the contrary, RLHB can realign the policy model immediately once human feedbacks are updated, owing to its multi-model joint training mechanism and natural language-style signal data modeling method.
Furthermore, online interaction process is often multi-turn and context-dependent.
Either pairwise-based reward models in RLHF or pointwise-based classifier models in RLHBC can only model one-round interaction, hindering scalable and sustainable training for online alignment.
However, the IRL mechanism behind RLHB and the natural language-based behavioral modeling provide room for solving this problem, which can also inspire future research.

% \section*{Acknowledgements}

% \bibliography{custom}
% \input{rlhb2024.bbl}
\bibliography{rlhb2024}

% \onecolumn
% \appendix
% \section{Appendix}
% \label{sec:appendix}

\end{document}